\documentclass[letterpaper]{article} 
\usepackage[preprint]{aaai2027}  
\usepackage[hyphens]{url}  
\usepackage{graphicx} 
\usepackage{natbib}  
\usepackage{caption} 
\usepackage{algorithm}
\usepackage{algorithmic}
\usepackage{amsmath}
\usepackage{amssymb}
\usepackage{mathtools}
\usepackage{amsthm}
\usepackage{tabularx}

\newtheorem{definition}{Definition}
\usepackage{newfloat}
\usepackage{listings}
\DeclareCaptionStyle{ruled}{labelfont=normalfont,labelsep=colon,strut=off} 
\floatstyle{ruled}
\newfloat{listing}{tb}{lst}{}
\floatname{listing}{Listing}

\usepackage{booktabs}

\title{When Errors Become Memories: Causal Pathway Tracing \\in Multi-Turn Memory-Augmented LLMs}
\author{
Shuyao Xiao\textsuperscript{\rm 1,\rm 2}
\thanks{This work was conducted during an internship at Beike Language and Intelligence.},
Shengling Wang\textsuperscript{\rm 1}\corresponding,
Xuan Chen\textsuperscript{\rm 2},
Ke Chao\textsuperscript{\rm 1},\\
Ming Cui\textsuperscript{\rm 2},
Feifei Qian\textsuperscript{\rm 1},
Fanlin Meng\textsuperscript{\rm 2},
Chaoyang Mei\textsuperscript{\rm 2},\\
Chaoyong Jiang\textsuperscript{\rm 1},
Qi Ouyang\textsuperscript{\rm 3},
Junxi Yin\textsuperscript{\rm 2},
}

\affiliations{
\textsuperscript{\rm 1}College of Artificial Intelligence, Beijing Normal University, Beijing, China\\
\textsuperscript{\rm 2}Beike Language and Intelligence, Beijing, China\\
\textsuperscript{\rm 3}Center for Data Science, New York University, New York, NY, USA\\
xiaoshuyao@mail.bnu.edu.cn, wangshengling@bnu.edu.cn,
chenxuan046@ke.com
}

\begin{document}

\maketitle

\begin{abstract}
Long-term memory enables large language models (LLMs) to preserve and reuse information across interactions, but it can also turn localized errors into persistent risks. Existing work mainly evaluates whether memory systems store and retrieve information correctly, leaving limited understanding of how errors propagate across responses, memory states, and future interactions. We propose a structural causal model (SCM)-based framework for cross-turn error propagation in memory-augmented LLMs. We model user questions, model responses, and memory states as a dynamic causal process, and identify two entry pathways: internal memory updating and external question feedback. By intervening on these pathways, we construct four counterfactual trajectories and quantify their downstream effects and interaction. Error influence is evaluated at four levels: memory retention, natural responses, targeted diagnostic probing, and probability-level error preference. Experiments show that error influence generally decays with interaction distance, while the memory-update pathway contributes more persistent effects than question feedback; latent errors may remain even after disappearing from natural responses. Propagation patterns also vary across memory categories and memory mechanisms. Pathway-guided restoration further validates this decomposition: Question Repair reduces residual error by 27.5\%, Memory Repair by 70.2\%, and Joint Repair by 98.3\%, nearly eliminating residual propagation.
\end{abstract}

\section{Introduction}

As large language models (LLMs) are increasingly deployed in long-term conversational settings, personalized assistants, and persistent interactive systems, external long-term memory has become an important mechanism for preserving and reusing information across sessions~\cite{zhang2025survey}. Memory systems such as MemoryBank~\cite{zhong2024memorybank}, Generative Agents~\cite{park2023generative}, and MemGPT~\cite{packer2023memgpt} enable models to move beyond the limitations of the context window through continuous memory updates, dynamic retrieval, and hierarchical memory management. Meanwhile, benchmarks such as LongMemEval~\cite{wu2024longmemeval}, MemBench~\cite{tan2025membench}, and MemoryAgentBench~\cite{hu2025evaluating} evaluate memory systems in terms of information retrieval, knowledge updating, long-range understanding, selective forgetting, and overall memory capability.

Long-term memory, however, also changes the nature of model errors. In conventional single-turn interactions, an incorrect response typically constitutes a localized output failure. In memory-augmented systems, by contrast, the same error may be incorporated into subsequent memory states and repeatedly retrieved and reused, thereby influencing responses across future interactions~\cite{xiong2025memory,zhang2023language,kwan2024mt}. In high-stakes settings that rely on persistent information, such as health management, financial planning, and task execution, these enduring effects may further shape user decisions or agent actions~\cite{dash2026untrustedinputtrustedmemory,qian-2026-visual}. Prior studies have shown that erroneous or contaminated information may persist once incorporated into long-term memory and continue to affect subsequent responses, plans, and behaviors when retrieved~\cite{xiong2025memory,deng2026memtrace}. Consequently, an initially isolated error may propagate over time, evolving into a long-term risk that affects both system states and real-world outcomes.

Existing research has primarily focused on the construction, management, and evaluation of long-term memory, typically assessing whether a model can correctly retain and use historical information through final-answer quality or aggregate task performance~\cite{wu2024longmemeval,tan2025membench,hu2025evaluating}. However, long-term memory is not just a static storage component; it is part of a dynamic system where user questions, model responses, and memory states evolve together. At the system level, an error may initially affect future behavior through both internal memory states and external interaction dynamics.  Internally, an erroneous response may be written into long-term memory, affecting subsequent generation~\cite{xiong2025memory,zhang2026useful,deng2026memtrace}. Externally, it may reshape the user's follow-up questions, causing the conversation to proceed from an incorrect premise~\cite{kwan2024mt}. These pathways may later interact, prolonging the influence of the original error. Existing approaches provide limited insight into how errors propagate dynamically across responses, memory states, and subsequent questions, and how the pathway through which an error enters the system shapes its downstream influence.

To address this gap, we introduce a structural causal model (SCM)-based framework~\cite{pearl2009causal} for analyzing cross-turn error propagation dynamics. The framework integrates the joint evolution of user questions, model responses, and memory states within a unified causal formulation and employs counterfactual interventions to control whether an error enters through internal memory, external question feedback, or both. We use \emph{error influence} to refer to the measurable downstream effects of an injected error on memory states and subsequent behavior, including memory retention, occurrence in natural responses, re-elicitation under targeted probes, and shifts in generation probabilities. By tracing these manifestations, our framework reveals how errors persist, decay, and re-emerge across long-term interactions, providing a unified perspective for understanding and mitigating systematic error propagation in memory-augmented LLMs. Our main contributions are as follows:

\begin{itemize}
    \item We develop a dynamic structural causal model and a four-trajectory counterfactual intervention framework that estimates the downstream effects of introducing an error through memory updating and question feedback, along with their interaction.

    \item Our experiments show that error influence generally decreases with interaction distance, while errors introduced through the internal memory pathway have more persistent downstream effects than those from the question-feedback pathway. Even after an error disappears from natural responses, it may remain as a latent memory trace or a shift in generation preference.

    \item Error propagation varies across memory categories and mechanisms, showing how memory organization shapes error persistence. We perform pathway-guided restoration to update the internal memory state and the external question-feedback trajectory. Restoring the corrupted memory state suppresses error propagation more effectively than correcting the error-affected question trajectory alone, while restoring both components nearly eliminates the residual effect.
\end{itemize}

\section{Related Work}

\paragraph{Long-Term Memory-Augmented LLMs.}
To support long-term dialogue, personalized interaction, and information reuse across sessions, prior studies have augmented large language models with external memory. MemoryBank~\cite{zhong2024memorybank} maintains continuously updated user memories and incorporates reinforcement and forgetting mechanisms based on recency and importance. Generative Agents~\cite{park2023generative} organize experiences into a natural-language memory stream and combine retrieval with reflection to support planning and temporally coherent behavior. MemGPT~\cite{packer2023memgpt} draws inspiration from virtual memory in operating systems and dynamically manages information between the active context and external storage, thereby enabling multi-session dialogue and long-document processing. LongMem~\cite{wang2023augmenting} adopts a decoupled architecture for memory encoding, retrieval, and reading, allowing language models to cache and utilize information from extended historical contexts.

These studies improve the storage, updating, retrieval, and management of historical information. Their primary objective, however, is to develop more effective memory architectures or improve downstream performance, with comparatively limited attention paid to how information propagates across multiple rounds of interaction.

\paragraph{Memory Evaluation in LLM Systems.}
As long-term memory mechanisms have evolved, recent work has begun to systematically evaluate memory capabilities in persistent interactions. LongMemEval~\cite{wu2024longmemeval} assesses whether conversational assistants can retrieve and integrate historical information through tasks involving information extraction, cross-session reasoning, temporal reasoning, knowledge updating, and abstention. MemBench~\cite{tan2025membench} evaluates memory systems along dimensions such as factual and reflective memory, participatory and observational interaction, effectiveness, efficiency, and capacity. MemoryAgentBench~\cite{hu2025evaluating} transforms long-context data into incremental, multi-turn interactions and evaluates four core capabilities: accurate retrieval, test-time learning, long-range understanding, and selective forgetting.

These benchmarks have advanced memory evaluation from static long-context question answering to incremental and interactive settings. However, they primarily evaluate a system's ability to retain, update, and utilize historical information, usually based on final-answer quality or aggregate task performance.

\paragraph{Causal Analysis of LLM Behavior.}
Causal intervention has become an important approach for understanding the generation mechanisms of large language models. Causal tracing perturbs and restores intermediate activations to identify internal computations that causally contribute to factual predictions~\cite{meng2022locating}. ROME~\cite{meng2022locating}, activation patching, and interchange interventions~\cite{zhang2023towards} further localize and manipulate specific layers, token positions, or model components, while causal mediation analysis decomposes input–output effects transmitted through selected intermediate representations~\cite{vig2020investigating}. Recent work has examined how the causal influence of input questions and generated prefixes evolves during autoregressive generation~\cite{xiao2026bridging}, but its analysis remains confined to a single generation trajectory.

Existing causal studies focus on internal variables within a single generation process, leaving cross-turn system dynamics underexplored. In memory-augmented LLMs, however, questions, responses, and memory states evolve jointly, allowing errors to persist and spread across multiple pathways. The contribution of these pathways to long-term error propagation remains poorly understood.

\section{Structural Causal Modeling}
\label{sec:scm}

\begin{figure}[t]
    \centering
    \includegraphics[width=1\linewidth]{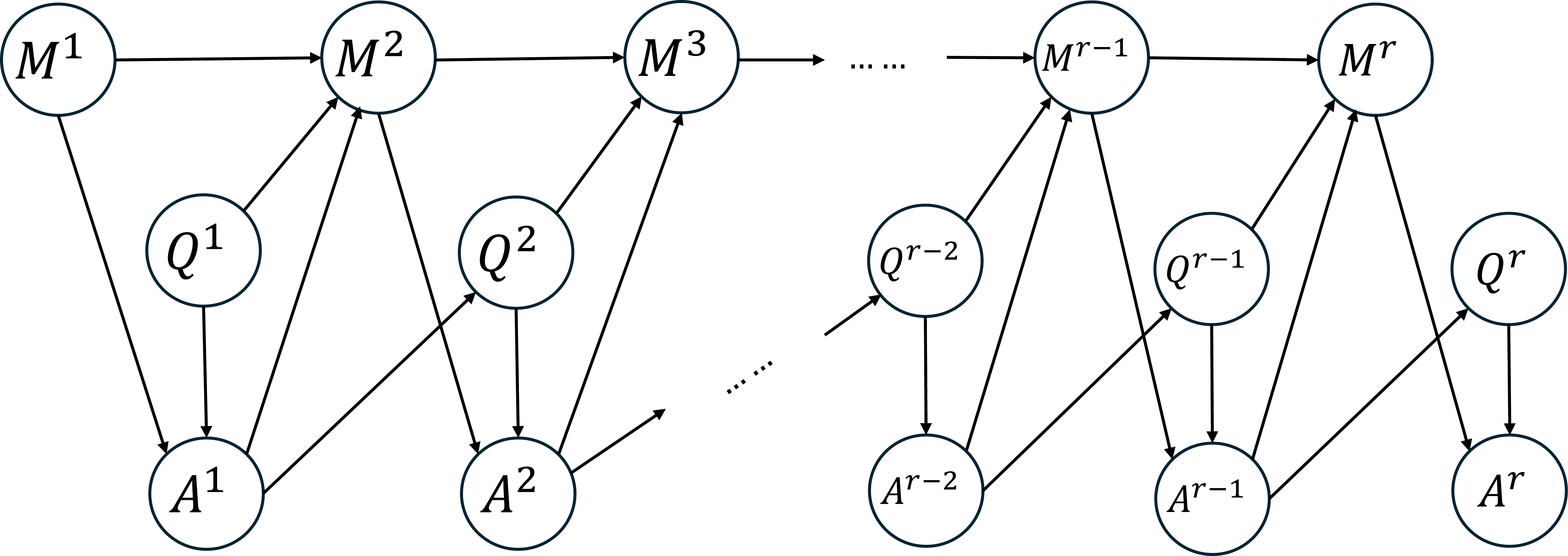}
    \caption{Dynamic structural causal graph of a long-term memory-augmented LLM agent. The memory-update path \(A^{r-1}\!\rightarrow M^r\!\rightarrow A^r\) and question-feedback path \(A^{r-1}\!\rightarrow Q^r\!\rightarrow A^r\) represent two mechanisms through which an erroneous response can affect subsequent interactions.}
    \label{fig:scm}
\end{figure}

To characterize the dynamic causal relationships among user questions, model responses, and memory updates in an LLM agent, we abstract the interaction and memory-update processes common to long-term memory systems and formulate the multi-turn interaction as a structural causal model indexed by interaction round~\cite{pearl2009causal}. As illustrated in Figure~\ref{fig:scm}, at round \(r\), \(Q^r\), \(A^r\), and \(M^r\) denote the user question, generated response, and memory state available before response generation, respectively.

We adopt a first-order Markov assumption to model the temporal evolution of the memory system. Variables at round \(r\) depend only on those from round \(r-1\) and the current round, while information from earlier rounds may still influence the current interaction through intermediate states. This assumption provides a tractable abstraction while preserving question evolution, response generation, and memory updating. Under this assumption, the causal graph includes four main types of relationships.

\paragraph{Memory persistence.}
The edge \(M^{r-1}\rightarrow M^r\) represents the persistence of
long-term memory across interaction rounds, capturing the retention
and transmission of stored information
~\cite{zhong2024memorybank,packer2023memgpt}.

\paragraph{Memory update.} The edges \(Q^{r-1}\rightarrow M^r\) and \(A^{r-1}\rightarrow M^r\) describe memory updating. The former indicates that user queries influence retained information, while the latter captures how model responses may be written into memory and become part of the future internal state~\cite{zhong2024memorybank,packer2023memgpt,xiong2025memory}.

\paragraph{Question continuation.} The edge \(A^{r-1}\rightarrow Q^r\) captures how the assistant's previous response influences the user's subsequent question. Users may pose follow-up questions, request corrections, or seek elaboration based on previous responses, causing the dialogue trajectory to depend on historical outputs~\cite{kwan2024mt,laban2025llms}.

\paragraph{Response generation.}
The edges \(Q^r\rightarrow A^r\) and \(M^r\rightarrow A^r\)
represent response generation, in which the current response is
conditioned jointly on the current question and the memory state
available at that round~\cite{zhong2024memorybank}.

Although the causal graph characterizes the normal evolution of a memory agent, an erroneous response may introduce incorrect information into subsequent interactions. Based on the system dynamics, we identify two primary pathways through which an erroneous response \(A^{r-1}\) can affect subsequent interactions: the memory-update pathway \(A^{r-1}\rightarrow M^r\rightarrow A^r\) and the question-feedback pathway \(A^{r-1}\rightarrow Q^r\rightarrow A^r\).

\paragraph{Memory-update path.} This path describes how erroneous information enters subsequent interactions through the agent's internal memory. Information from an erroneous response may be extracted and stored by the memory module through \(A^{r-1}\rightarrow M^r\). Because \(A^r\) depends on \(M^r\), the corrupted memory may bias generation toward the same error~\cite{xiong2025memory,zhang2026useful,deng2026memtrace}.

\paragraph{Question-feedback path.} This path describes how an error enters subsequent interactions by shaping the user's next question. An erroneous response may affect the next question through \(A^{r-1}\rightarrow Q^r\). For example, a user may continue the conversation based on an incorrect premise introduced by the model, causing subsequent responses to follow a trajectory shaped by the original error~\cite{laban2025llms}.

The two entry pathways affect different components of the causal system. The memory-update pathway changes the agent's internal state, whereas the question-feedback pathway changes the external interaction context. Over long-term interactions, these mechanisms may interact: an erroneous response may alter future questions, trigger additional errors, and cause further incorrect information to be written into memory. Thus, the two pathways represent distinct mechanisms through which an error enters the system at the intervention turn and propagates across subsequent interactions.

To quantify the downstream effects of each entry mechanism, we introduce controlled causal interventions in the following section.

\section{Causal Intervention and Pathway Effects}
\label{sec:intervention}

Let \(t\) denote the error-injection round. Following the two outgoing causal edges from the injected response \(A^t\) to the next-round memory state \(M^{t+1}\) and user question \(Q^{t+1}\) in Figure~\ref{fig:scm}, we intervene on the inputs to the memory updater and the next-question generator. These two successor mechanisms define a \(2\times2\) joint intervention design.

At round \(t\), we replace the naturally generated response with a matched correct response \(A_c^t\) or erroneous response \(A_e^t\) to the same question. The two responses differ only in the target proposition, while the remaining content and wording are preserved as closely as possible. Passing them separately to the memory updater produces a clean memory state \(M_c^{t+1}\) and an erroneous memory state \(M_e^{t+1}\), respectively. Combining the correct and erroneous response variants as inputs to the two mechanisms yields four counterfactual trajectories~\cite{shpitser2014causal}:
\begin{equation}
\begin{aligned}
\mathcal{T}_{cc} &= \left(M_c^{t+1}, A_c^t\right), \qquad
\mathcal{T}_{ec} = \left(M_e^{t+1}, A_c^t\right),\\
\mathcal{T}_{ce} &= \left(M_c^{t+1}, A_e^t\right), \qquad
\mathcal{T}_{ee} = \left(M_e^{t+1}, A_e^t\right).
\end{aligned}
\end{equation}

The first component specifies the memory state available at round \(t+1\), while the second specifies the response supplied to the next-question generator. Accordingly, the first subscript denotes the intervention on the memory-update mechanism, and the second denotes the intervention on the question-feedback mechanism; \(c\) and \(e\) indicate correct and erroneous inputs, respectively. Thus, \(\mathcal{T}_{cc}\) is the fully clean trajectory, \(\mathcal{T}_{ec}\) introduces the error through memory updating only, \(\mathcal{T}_{ce}\) introduces it through question feedback only, and \(\mathcal{T}_{ee}\) introduces it through both mechanisms.

After the intervention at round \(t\), the four trajectories evolve independently according to their respective memory states, dialogue histories, and previous responses. Their differences therefore identify the downstream causal contrasts induced by the assigned error-entry interventions, while subsequent memory updating, question evolution, and response generation unfold naturally within each trajectory.

\subsection{Error-Propagation Outcomes}
\label{sec:error_outcomes}

Let \(k\) denote the interaction distance from the error-injection round, where \(k=1\) corresponds to the first post-intervention round. We measure the downstream influence of the injected error at four complementary levels: memory retention, natural responses, targeted diagnostic probing, and generation preference. Within a given trajectory, let \(M(k)\) denote the memory state available immediately before response generation at interaction distance \(k\). The memory-state outcome is defined as
\begin{equation}
Y_M(k)=\mathbb{I}\left[\text{the target error is detected in }M(k)\right],
\end{equation}
where \(Y_M(k)=1\) indicates that the target error remains in the current memory state. The natural-response outcome is defined as
\begin{equation}
Y_A(k)=\mathbb{I}\left[\text{the natural response contains the target error}\right].
\end{equation}
Here, \(Y_A(k)=1\) when the response explicitly states, necessarily assumes, or directly relies on the target error.

Because natural follow-up questions may not directly concern the target fact, we also use a targeted diagnostic question \(q_k\) that asks about it directly. At each interaction distance \(k\), the same probe \(q_k\) is used across all four trajectories to ensure comparability. The controlled-probe outcome is defined as
\begin{equation}
Y_C(k)=\mathbb{I}\left[\text{the response to }q_k\text{ contains the target error}\right],
\end{equation}
where \(Y_C(k)=1\) indicates that the target error can be directly elicited from the current trajectory state. The outcomes \(Y_M(k)\), \(Y_A(k)\), and \(Y_C(k)\) are determined by a proposition-level evaluator. A text is labeled erroneous only if it explicitly states, necessarily assumes, or directly relies on the complete target error; merely quoting, negating, or correcting the erroneous proposition does not count as propagation.

To capture latent shifts that may not appear in the final decoded response, we construct a matched erroneous target answer \(a_{\mathrm{err}}\) and correct target answer \(a_{\mathrm{cor}}\) for \(q_k\), and define the probability-level error preference as
\begin{equation}
\resizebox{0.90\columnwidth}{!}{$
Y_P(k)=\frac{1}{\ln 2}\left[
\frac{\log P(a_{\mathrm{err}}\mid q_k,M(k))}{\lvert a_{\mathrm{err}}\rvert}
-
\frac{\log P(a_{\mathrm{cor}}\mid q_k,M(k))}{\lvert a_{\mathrm{cor}}\rvert}
\right],
$}
\end{equation}
where \(\lvert a\rvert\) denotes the number of tokens in answer \(a\), and the factor \(1/\ln 2\) converts the difference in natural log-likelihoods into bits. A positive \(Y_P(k)\) indicates a preference for the erroneous target, whereas a negative value indicates a preference for the correct target. The four outcomes capture, respectively, error retention in memory, explicit error occurrence in natural responses, error re-elicitation through targeted probing, and latent shifts in generation preference.

\begin{figure*}[t]
    \centering
    \includegraphics[width=1\textwidth]{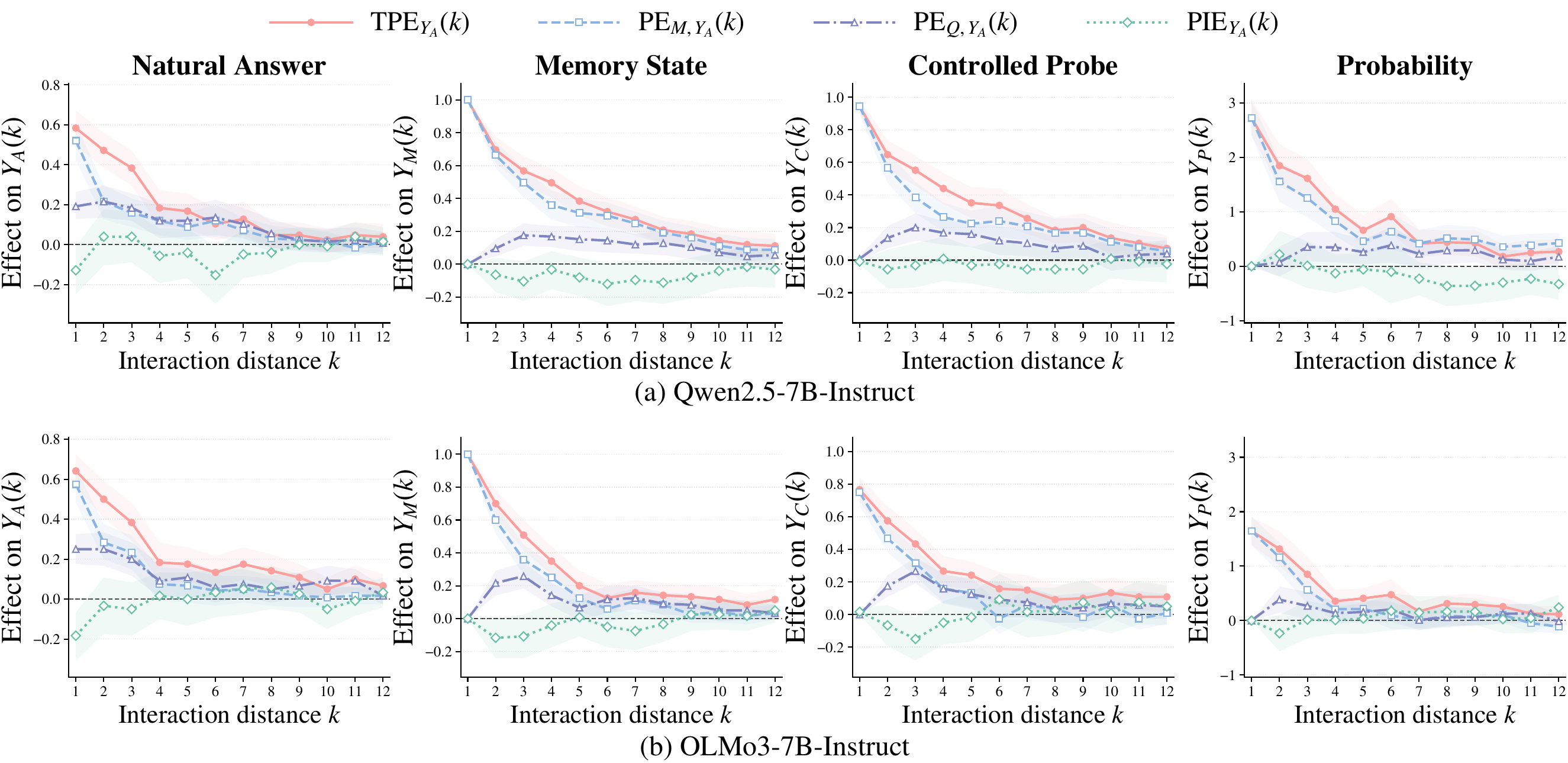}
    \caption{Pathway effects of an injected error across interaction distances for two models. The first and second rows show results for Qwen2.5-7B-Instruct and OLMo3-7B-Instruct, respectively. From left to right, the columns correspond to \(Y_A(k)\), \(Y_M(k)\), \(Y_C(k)\), and \(Y_P(k)\). The curves show the total propagation effect \(\mathrm{TPE}\), memory-update pathway effect \(\mathrm{PE}_M\), question-feedback pathway effect \(\mathrm{PE}_Q\), and pathway interaction effect \(\mathrm{PIE}\); shaded regions indicate 95\% bootstrap confidence intervals.}
    \label{fig:path_effects_model_comparison}
\end{figure*}

\subsection{Pathway Effects}
\label{sec:effect_decomposition}

Motivated by causal contrasts for pathway and joint interventions~\cite{avin2005identifiability,shpitser2014causal}, we define trajectory-level effects according to the mechanism through which the error initially enters the system. For any outcome \(Y\in\{Y_M,Y_A,Y_C,Y_P\}\), let \(Y_{xz}(k)\) denote its value under trajectory \(\mathcal{T}_{xz}\) at interaction distance \(k\), where \(x,z\in\{c,e\}\), and let \(\mathbb{E}[\cdot]\) denote the average over experimental instances.

\begin{definition}[\textbf{Total Propagation Effect}]
The total propagation effect at interaction distance \(k\) is defined as
\label{def:total_effect}
\begin{equation}
\mathrm{TPE}_{Y}(k)=\mathbb{E}[Y_{ee}(k)]-\mathbb{E}[Y_{cc}(k)].
\end{equation}
\end{definition}
It measures the overall downstream effect of introducing the error through both entry mechanisms relative to the fully clean trajectory.

\begin{definition}[\textbf{Memory-Update Pathway Effect}]
\label{def:memory_entry_effect}
The memory-update pathway effect at interaction distance \(k\) is defined as
\begin{equation}
\mathrm{PE}_{M,Y}(k)=\mathbb{E}[Y_{ec}(k)]-\mathbb{E}[Y_{cc}(k)].
\end{equation}
\end{definition}
It measures the downstream effect associated with introducing the error only through the memory updater.

\begin{definition}[\textbf{Question-Feedback Pathway Effect}] The question-feedback pathway effect at interaction distance \(k\) is defined as
\label{def:question_entry_effect}
\begin{equation}
\mathrm{PE}_{Q,Y}(k)=\mathbb{E}[Y_{ce}(k)]-\mathbb{E}[Y_{cc}(k)].
\end{equation}
\end{definition}
It measures the downstream effect associated with introducing the error only through the next-question generator.

\begin{definition}[\textbf{Pathway Interaction Effect}] The pathway interaction effect is defined as
\label{def:interaction_effect}
\begingroup
\small
\begin{equation}
\begin{aligned}
\mathrm{PIE}_{Y}(k)
&= \mathrm{TPE}_{Y}(k)
 - \mathrm{PE}_{M,Y}(k)
 - \mathrm{PE}_{Q,Y}(k) \\
&= \mathbb{E}[Y_{ee}(k)]
 - \mathbb{E}[Y_{ec}(k)]
 - \mathbb{E}[Y_{ce}(k)]
 + \mathbb{E}[Y_{cc}(k)].
\end{aligned}
\end{equation}
\endgroup

\end{definition}

These estimands capture the downstream effects associated with the pathway through which the error is introduced at the injection turn, while each trajectory subsequently evolves according to its own states and dialogue history. The interaction effect measures non-additivity between the two pathway interventions, with positive and negative values indicating super-additive and sub-additive effects, respectively.
For example, \(\mathrm{TPE}_{Y_M}(k)\), \(\mathrm{PE}_{M,Y_A}(k)\), and \(\mathrm{PE}_{Q,Y_C}(k)\) denote the total effect on memory retention, the memory-update pathway effect on natural responses, and the question-feedback pathway effect on controlled elicitation, respectively. All trajectories share the same pre-intervention history, and the same diagnostic question is used across all four trajectories for
controlled-probe and probability-level evaluation.


\section{Experimental Evaluation}
\label{sec:experiments}
\paragraph{Dataset Construction.} 
Existing multi-turn dialogue datasets typically use fixed interaction trajectories and do not support the pathway-specific interventions required by our causal design. Therefore, we construct 144 multi-turn interactions across six memory categories, with 24 instances each. Each interaction contains 15 turns; one target error is injected at turn 3 and traced over the following 12 turns, where \(k\in\{1,\ldots,12\}\) denotes the interaction distance. Construction details are provided in the supplementary material.

\paragraph{Memory Mechanisms and Models.}
The main setting follows a MemoryBank-style mechanism~\cite{zhong2024memorybank} where the memory state is updated after each interaction round. We additionally evaluate an episodic-retrieval variant inspired by prior retrieval-based memory systems~\cite{park2023generative,packer2023memgpt}. This variant stores each interaction as an independent memory entry and retrieves the three most relevant entries using a deterministic, length-normalized lexical-overlap score, with recency used to break ties. We use Qwen2.5-7B-Instruct~\cite{qwen2025qwen25technicalreport} and OLMo3-7B-Instruct~\cite{olmo2026olmo3} as answer models, while Qwen2.5-32B-Instruct~\cite{qwen2025qwen25technicalreport} is used for question simulation, counterfactual construction, and proposition-level evaluation.

\paragraph{Experimental Settings.}
Natural responses, memory updates, controlled-probe questions and
responses, and proposition-level evaluator outputs use greedy decoding. Subsequent questions are sampled with temperature \(0.7\) and top-\(p=0.9\); failed answer-pair validations are retried with temperature \(0.7\) and top-\(p=0.95\), for at most five attempts. Maximum generation lengths are 96, 128, 256, 256, and 192 tokens for questions, answers, memory updates, answer pairs, and evaluator outputs, respectively.

\paragraph{Valid Instances and Statistical Reporting.}
We retain instances where the correct response yields a clean memory state and the erroneous response successfully writes the target error, resulting in 125 valid instances for Qwen2.5-7B-Instruct and 120 for OLMo3-7B-Instruct. Each valid instance is executed under each of the four counterfactual trajectories, and reported effects are averaged over instances. Shaded regions denote 95\% percentile bootstrap confidence intervals computed from 5,000 instance-level resamples. All experiments use 8 NVIDIA H200 GPUs.

\begin{figure*}[t]
    \centering
    \includegraphics[width=\textwidth]{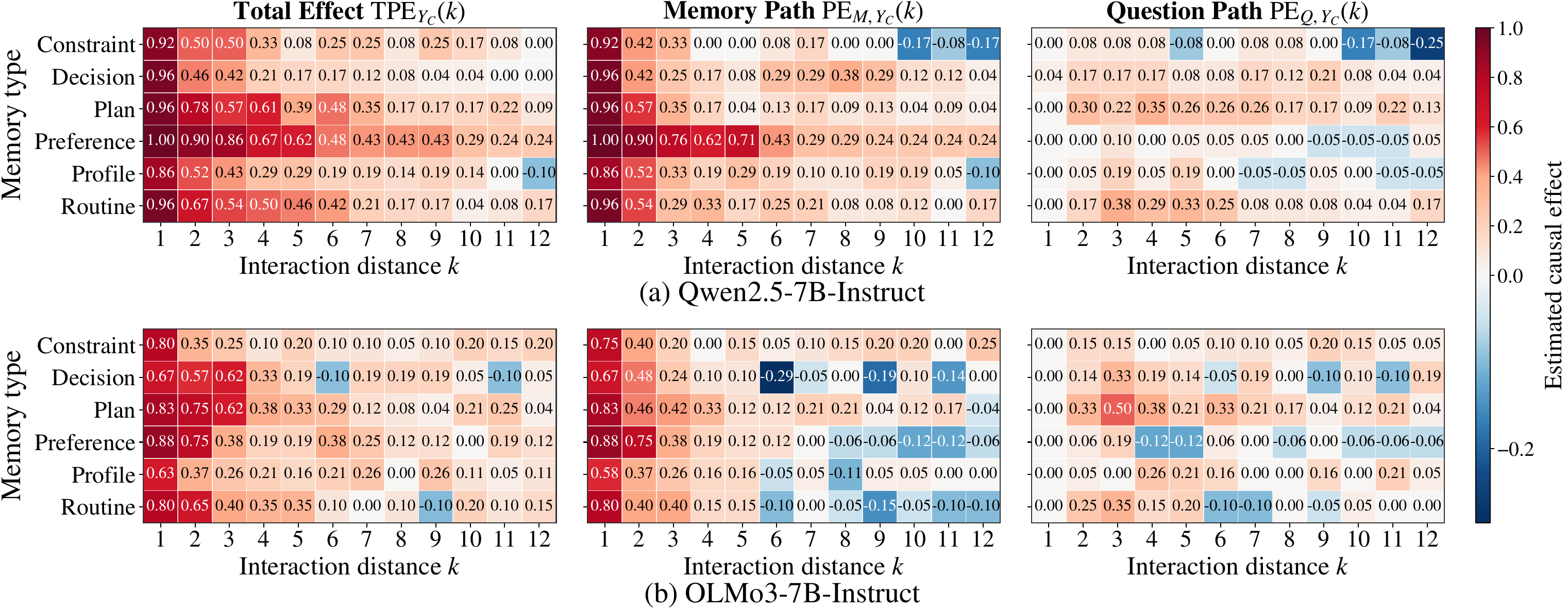}
    \caption{Error propagation across six memory categories for (a) Qwen2.5-7B-Instruct and (b) OLMo3-7B-Instruct. From left to right, the columns show the total propagation effect \(\mathrm{TPE}_{Y_C}(k)\), memory-update pathway effect \(\mathrm{PE}_{M,Y_C}(k)\), and question-feedback pathway effect \(\mathrm{PE}_{Q,Y_C}(k)\). Colors indicate effect magnitude under a shared scale.}
    \label{fig:memory_type_effects}
\end{figure*}

\subsection{Error Propagation and Pathway Effects}
\label{sec:path_effect_results}

We evaluate error propagation using four complementary outcomes defined in Section~\ref{sec:error_outcomes}: the natural-response outcome \(Y_A\), memory-state outcome \(Y_M\), controlled-probe outcome \(Y_C\), and probability-level error preference \(Y_P\). To assess cross-model robustness, we repeat the experiment with Qwen2.5-7B-Instruct and OLMo3-7B-Instruct under identical data, memory, intervention, and evaluation settings.

As shown in Figure~\ref{fig:path_effects_model_comparison}, both models exhibit strong short-range propagation followed by gradual decay across all four outcomes. The natural-response effect decreases most rapidly: for Qwen2.5-7B-Instruct and OLMo3-7B-Instruct, the total effect falls from \(0.584\) and \(0.642\) at \(k=1\) to \(0.040\) and \(0.067\) at \(k=12\), respectively. In contrast, memory-state and controlled-probe outcomes reveal longer-lasting traces. At \(k=1\), the memory-state total effect reaches \(1.0\) for both models, while the controlled-probe effect reaches \(0.944\) and \(0.767\), showing that, among valid injections, the error remains present in the next-round memory state and can be reliably re-elicited.

The probability-level outcome provides a more fine-grained view of latent propagation. At \(k=1\), the total effects are \(2.726\) for Qwen2.5-7B-Instruct and \(1.646\) for OLMo3-7B-Instruct, indicating a substantial shift in relative preference toward the erroneous target. Thus, an error may disappear from natural responses while remaining stored in memory, recoverable under targeted probing, or reflected in the underlying generation distribution.

The pathway analysis shows that the memory-update pathway contributes more strongly to \(Y_M\), \(Y_C\), and \(Y_P\). The question-feedback pathway is generally weaker and relies more on whether later questions reactivate error-related information. The interaction effects are mostly negative or near zero. This may happen because the two pathways have overlapping downstream influences: once one pathway induces the error, redundancy or outcome
saturation limits the effect of activating the other. These patterns hold across both answer models, though controlled-probe and probability-level traces generally last longer for Qwen2.5-7B-Instruct, while OLMo3-7B-Instruct shows faster attenuation.

\begin{table}[t]
    \centering
    \small
    \setlength{\tabcolsep}{3.8pt}
    \begin{tabular}{lccc}
        \toprule
        Comparison & Evaluation Set & Agreement & \(\kappa\) \\
        \midrule
        Qwen--GLM & Full (\(35{,}280\))       & 91.9\% & 0.709 \\
        Qwen--GPT & Stratified (\(900\))      & 89.3\% & 0.591 \\
        GLM--GPT  & Stratified (\(900\))      & 96.1\% & 0.805 \\
        \bottomrule
    \end{tabular}
    \caption{Binary agreement among proposition-level evaluators.
    GPT results are post-stratified to the full experiment.}
    \label{tab:judge_robustness}
\end{table}
\paragraph{Evaluator robustness.}
We assess judge dependence by re-evaluating all \(35{,}280\)
proposition-level outcomes with GLM-5.2~\cite{glm5team2026glm5vibecodingagentic} and a stratified sample of \(900\) outcomes with GPT-5.5~\cite{openai2026gpt55}. As shown in Table~\ref{tab:judge_robustness}, Qwen agrees strongly with both external judges. Although GLM yields lower absolute error rates than the primary evaluator, re-judging preserves the direction of all 72 \(\mathrm{TPE}\) estimates and yields temporal-curve correlations of \(r=.975\)--\(.996\). The stronger contribution of the memory-update pathway and the negative or near-zero interactions also remain unchanged. Thus, evaluator calibration affects effect magnitude but not our temporal or pathway-level conclusions.

\subsection{Propagation Patterns across Memory Types}
\label{sec:memory_type_analysis}

We further compare error propagation across six memory categories: \textit{Constraint}, \textit{Decision}, \textit{Plan}, \textit{Preference}, \textit{Profile}, and \textit{Routine}. To reduce variation caused by natural follow-up questions, we use a fixed diagnostic question that directly probes the target fact.

As shown in Figure~\ref{fig:memory_type_effects}, propagation varies across memory categories and answer models. For Qwen2.5-7B-Instruct, preference errors are the most persistent and receive larger contributions from the memory-update pathway, while plan and routine errors receive greater contributions from question feedback. For OLMo3-7B-Instruct, plan errors are the most persistent and are sustained by both memory retention and changes in subsequent questions.
Across both models, preference errors are primarily associated with persistent memory retention, while plan errors significantly alter later interaction trajectories. Routine errors receive contributions from both pathways, while constraint, decision, and profile errors generally decay faster or fluctuate more across rounds. Overall, memory category affects both the duration and mechanism of error propagation, and cross-model differences indicate that these patterns also depend on how each model stores and utilizes long-term information.

\subsection{Propagation Patterns across Memory Mechanisms}
\label{sec:memory_mechanism_comparison}

To examine whether the identified propagation patterns generalize across memory mechanisms, we compare two long-term memory architectures: continuously updated MemoryBank~\cite{zhong2024memorybank} and episodic retrieval with independent entries and Top-3 retrieval~\cite{park2023generative,packer2023memgpt}. MemoryBank consolidates historical interactions into a dynamic memory state, whereas episodic retrieval preserves interactions as independent entries and retrieves those most relevant to the current query.

\begin{figure}[t]
    \centering
    \includegraphics[width=0.9\linewidth]{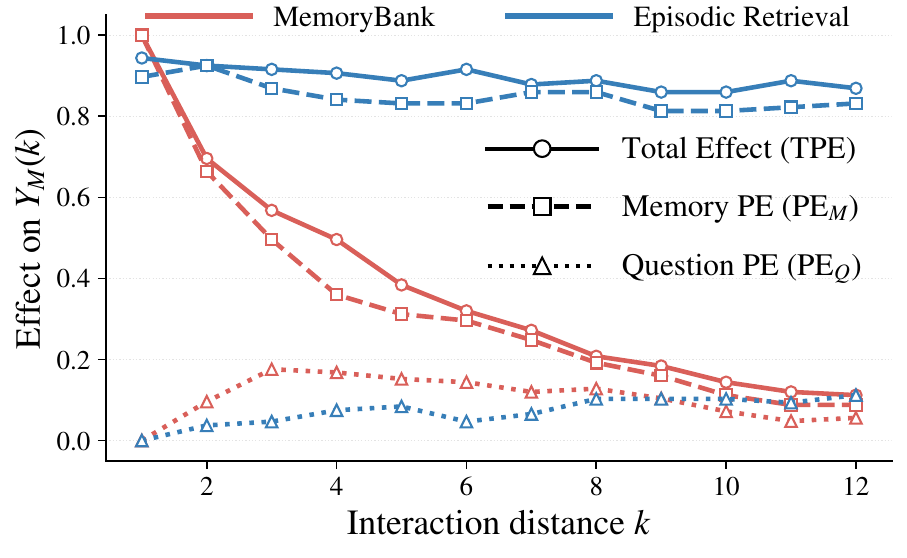}
    \caption{Effects of the injected error on the memory-state outcome \(Y_M(k)\) under different memory mechanisms. Red and blue curves denote MemoryBank and episodic retrieval, respectively. Line styles indicate the total propagation effect \(\mathrm{TPE}\), memory-update pathway effect \(\mathrm{PE}_M\), and question-feedback pathway effect \(\mathrm{PE}_Q\).}\label{fig:episodic_memory}
\end{figure}

As shown in Figure~\ref{fig:episodic_memory}, the two mechanisms exhibit different error-persistence patterns. Under MemoryBank, both \(\mathrm{TPE}\) and \(\mathrm{PE}_M\) decrease steadily as the memory state is updated. In contrast, episodic retrieval retains the erroneous entry in memory, resulting in persistently high memory-state effects over long interaction distances. 
Despite the difference in persistence, both mechanisms exhibit the same pathway-level pattern: \(\mathrm{PE}_M\) accounts for most of the memory-state effect, whereas \(\mathrm{PE}_Q\) remains comparatively small. Thus, memory-state error persistence depends both on the error-entry pathway and the subsequent memory mechanism.

\subsection{Can Pathway Effects Guide Error Restoration?}
\label{sec:path_restoration}

To examine whether pathway effects can guide effective interventions, we conduct a pathway-guided state-restoration experiment. The error is injected at turn 3, and a one-time restoration is applied before the turn at interaction distance \(k=3\). We then evaluate the remaining propagation over subsequent interactions. We conduct this experiment using Qwen2.5-7B-Instruct with the MemoryBank-style mechanism. We consider four strategies: \emph{No Repair}, \emph{Question Repair}, \emph{Memory Repair}, and \emph{Joint Repair}. \emph{Question Repair} restores the dialogue history and previous response while retaining the error-affected memory. \emph{Memory Repair} restores the memory state while retaining the error-affected dialogue context. \emph{Joint Repair} restores both components.

Using fixed controlled probes, we define the residual effect of strategy \(s\) at post-restoration distance \(j\) as $R_s(j)=Y_s(j)-Y_{\mathrm{clean}}(j)$,
where \(Y_s(j)\) and \(Y_{\mathrm{clean}}(j)\) denote the controlled-probe outcomes under strategy \(s\) and the clean trajectory. We summarize the remaining propagation using the mean residual effect:
\[
\operatorname{MRE}(s)=\frac{1}{J}\sum_{j=0}^{J-1}R_s(j),
\]
where \(J\) is the number of post-restoration distances. Lower MRE indicates more effective restoration. We also report the relative reduction compared with \emph{No Repair}:

\[
\operatorname{Reduction}(s)=\frac{\operatorname{MRE}(\mathrm{NoRepair})-\operatorname{MRE}(s)}{\operatorname{MRE}(\mathrm{NoRepair})}\times 100\%.
\]

\begin{table}[t]
    \centering
    \begin{tabular}{lcc}
        \toprule
        \textbf{Strategy} & \textbf{Residual MRE} $\downarrow$ & \textbf{Reduction} $\uparrow$ \\
        \midrule
        \textit{No Repair} & 0.178 & -- \\
        \textit{Question Repair} & 0.129 & 27.5\% \\
        \textit{Memory Repair} & 0.053 & 70.2\% \\
        \textit{Joint Repair} & \textbf{0.003} & \textbf{98.3\%} \\
        \bottomrule
    \end{tabular}
    \caption{Pathway-guided state-restoration results on controlled probes. Lower mean residual effect (MRE) indicates more effective restoration.}
    \label{tab:path_restoration}
\end{table}

As shown in Table~\ref{tab:path_restoration}, \emph{Memory Repair} reduces residual MRE by 70.2\%, compared with 27.5\% for \emph{Question Repair}, consistent with the stronger memory-update pathway effects identified in our causal analysis. On held-out instances, larger question-feedback pathway effects are associated with greater gains from \emph{Question Repair} on controlled probes (\(\rho=0.335\), 95\% CI \([0.158,0.493]\)). \emph{Joint Repair} further reduces residual propagation by 98.3\%, demonstrating that coordinated correction of the memory state and dialogue context is most effective. Overall, pathway effects quantify error propagation and indicate which components are more likely to benefit from intervention.
\section{Conclusion}

We present a dynamic structural causal framework for characterizing cross-turn error propagation in long-term memory-augmented LLMs and decomposing its total effect into memory-update, question-feedback, and interaction components. Experiments across models, memory categories, and memory mechanisms show that errors can persist across multiple turns, with the memory-update pathway typically contributing more strongly and latent errors remaining beyond observable responses. Pathway-guided restoration further shows that repairing the memory-update pathway is more effective than repairing the error-affected question-feedback pathway alone, while jointly repairing both pathways nearly eliminates residual propagation.

\appendix

\bibliography{aaai2027}


\end{document}